\documentclass{article} % For LaTeX2e
\usepackage{iclr2023_conference,times}

\usepackage[utf8]{inputenc} % allow utf-8 input
\usepackage[T1]{fontenc}    % use 8-bit T1 fonts
\usepackage{hyperref}       % hyperlinks
\usepackage{url}            % simple URL typesetting
\usepackage{booktabs}       % professional-quality tables
\usepackage{amsfonts}       % blackboard math symbols
\usepackage{amsthm}
\usepackage{nicefrac}       % compact symbols for 1/2, etc.
\usepackage{microtype}      % microtypography
\usepackage{xcolor}         % colors
\usepackage{wrapfig}
\usepackage{bbding}

\usepackage[pdftex]{graphicx}
\usepackage{caption}
\usepackage{subcaption}
\usepackage{hyperref}
\hypersetup{
    colorlinks=true,
    citecolor=black,
    linkcolor=black,
    urlcolor=blue,
    }
\usepackage{algorithm}
\usepackage{algpseudocode}
\usepackage{amsmath}
\usepackage{mathtools}
\usepackage{ragged2e}

\usepackage{xspace}

\usepackage{textgreek}
\usepackage[normalem]{ulem}

\theoremstyle{definition}
\newtheorem{definition}{Definition}[section]
\newtheorem{theorem}{Theorem}[section]

\allowdisplaybreaks[1]
\mathchardef\mhyphen="2D

\title{Composable Function-preserving Expansions for Transformer Architectures}
\author{
Andrea Gesmundo$^1$, \ Kaitlin Maile$^{1,2}$ \\
    \ ${ }^1$ Google DeepMind, ${ }^2$ IRIT, University of Toulouse, \\
    \texttt{\{agesmundo,kmaile\}@google.com} \\
}

\begin{document}

\maketitle

\begin{abstract}

Training state-of-the-art neural networks requires a high cost in terms of compute and time. Model scale is recognized to be a critical factor to achieve and improve the state-of-the-art. Increasing the scale of a neural network normally requires restarting from scratch by randomly initializing all the parameters of the model, as this implies a change of architecture's parameters that does not allow for a straightforward transfer of knowledge from smaller size models.

In this work, we propose six composable transformations to incrementally increase the size of transformer-based neural networks while preserving functionality, allowing to expand the capacity of the model as needed. We provide proof of exact function preservation under minimal initialization constraints for each transformation. The proposed methods may enable efficient training pipelines for larger and more powerful models by progressively expanding the architecture throughout training. \footnote{Implementation of the proposed transformations and empirical tests of the function preservation property are available at: \url{http://goo.gle/TransformerExpansions}.}

%\todo{(Ideally)} The experiments reported demonstrate that it is possible to (at least) match the quality achieved of a larger scale NN trained from random initialization by starting from a pretrained smaller NN that is incrementally extended by applying the proposed methods incurring in a lower training cost while resulting in a matching architecture with equal or better quality performance.

\end{abstract}

\section{Introduction}

%High level context}
Transformer-based neural networks have gained widespread attention in recent years due to their impressive performance. The Transformer architecture, introduced by \citet{Vaswani2017AttentionIA}, has become the standard for many natural language processing (NLP) tasks, including machine translation, text generation, and question answering. The success of transformer-based models is not limited to NLP: they have also been applied to various other domains, including computer vision, speech recognition, and recommendation systems. The largest and most performant of these models, large language models (LLMs) and vision and multimodal foundation models, are reaching billions to trillions of parameters \citep{Dehghani2023ScalingVT, touvron2023llama, Rae2021ScalingLM, Raffel2020ExploringTL}.

%Scaling challenges}
However, each new model is generally trained from scratch, without reusing the capabilities acquired by previously trained smaller models.
Furthermore, the size of the model is constant throughout training. The computational cost of training scales quadratically with model size due to the necessary increase in amount of training data \citep{hoffmann2022training, google2023palm, Kaplan2020ScalingLF}. The ability to reuse parameters of a pretrained model or dynamically increase a model's size during training could thus reduce the overall cost of training, but how to accomplish parameter reuse effectively without losing training progress is not straightforward.
% Simply adding randomly initialized parameters to an existing model would catastrophically change the model's functionality and performance, while adding only zero-initialized parameters wouldn't permit the necessary gradient flow to change these parameters away from zero and use the newly added capacity.

%We propose ..}
To address these limitations, we propose parameter expansion transformations for transformer-based models that are exactly function preserving. These transformations increase the model size and thus the potential capacity of the model without changing its functionality, permitting continued training. These composable transformations operate on independent dimensions of the architecture, allowing for fine-grained architectural expansion. 

Some previous works have also proposed function preserving parameter expansion transformations for transformer-based models \citep{chen2022bert2bert, shen2022staged, wang2023learning, mazzawi2023deep}, extending from techniques for smaller convolutional and dense models \citep{Chen2016Net2NetAL, Evci2022GradMaxGN}.
Our framework is so far the most comprehensive and composable set of function preserving transformations. 

%The contribution of this paper ...}
The contributions of this paper are six composable function preserving transformations applicable to Transformer architectures:
1) size of MLP internal representation,
2) number of attention heads,
3) size of the attention heads output representation,
4) size of the attention input representation,
5) size of the transformer layers input/output representations, 
6) number of layers,
summarized in Table \ref{table:summary1}.
For each transformation,
we provide proof of how the \emph{exactly function preserving} property is achieved with a minimal set of constraints on the initialization of the added parameters.

\begin{figure}[t]
  \centering
%   \fbox{\rule[-.5cm]{0cm}{4cm} \rule[-.5cm]{4cm}{0cm}}
%   \vspace{-40pt}
%   \hspace*{-40.7pt}
  \includegraphics[width=0.4\linewidth]{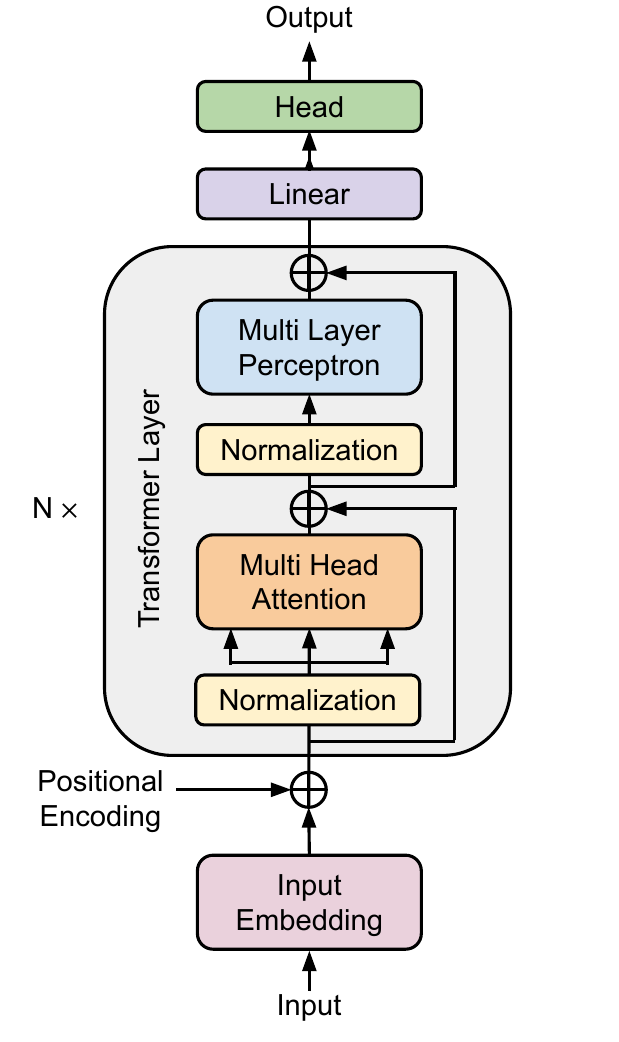}
  \caption{Representation of a standard Neural Network based on the Transformer architecture.
  \label{architecture}
}
\end{figure}

\section{Transformer architecture formalization}
\label{sec:trns}

This presentation is based on a particular instantiation of the transformer architecture: applications to variants (e.g. Encoder+Decoder, different normalization placement) can be obtained with simple extensions. 

Figure \ref{architecture} represents the standard Transformer architecture \citep{Vaswani2017AttentionIA}.
The \emph{Input Embedding} module maps the arbitrary input modality (e.g. image, text) into a bidimensional tensor $\underset{s\times h}{\mathrm{I}}$, where $s$ is the sequence dimension and $h$ is the hidden dimension.
The $\mathrm{TransformerArchitecture}(\cdot)$ is defined as a function that maps:  $\underset{s\times h}{\mathrm{I}} \rightarrow \underset{s\times o}{\mathrm{O}}$, where $o$ is the hidden dimension of the output representation.
The \emph{Head} component represents the output modality specific logic that maps $\underset{s\times o}{\mathrm{O}}$ into a specific output (e.g. a distribution over classes or text tokens).

$\mathrm{TransformerArchitecture}(\cdot)$ is defined as:

\begin{equation}
\label{eq:layers}
\mathrm{TransformerArchitecture}(\underset{s\times h}{\mathrm{I}}) = \mathrm{TransformerLayer}^{\circ N}(\underset{s\times h}{\mathrm{I}}\!+\!\underset{s\times h}{\mathbf{P}})\ \times \underset{h\times o}{\mathbf{W}^{out}},
\end{equation}

where
$\underset{h\times o}{\mathbf{W}^{out}}$ are the parameters of the final linear projection, 
$\underset{s\times h}{\mathbf{P}}$ are the positional embedding parameters,
and $\mathrm{TransformerLayer}^{\circ N}(\cdot)$ represents the recursive application of $N$ transformer layers.
The $n^\text{th}$ transformer layer is defined as:
\begin{equation}
\begin{array}{l}
\mathrm{TransformerLayer}_n(\underset{s\times h}{\mathrm{I}_n}) = \underset{s\times h}{\mathrm{I^{'}}_n} + \mathrm{MLP}_n(\mathrm{Norm}_n^\mathrm{MLP}(\underset{s\times h}{\mathrm{I^{'}}_n})), \\
\underset{s\times h}{\mathrm{I^{'}}_n} = \underset{s\times h}{\mathrm{I}_n} + \mathrm{MHA}_n(\mathrm{Norm}_n^\mathrm{MHA}(\underset{s\times h}{\mathrm{I}_n})) \\
\end{array}
\begin{array}{r}
\ \ \ \ \forall \ \ n \in [1, N].
\end{array}
\end{equation}

$\mathrm{MLP}_n(\cdot)$ is the \emph{Multi Layer Perceptron} (i.e. feed forward layers), defined as:

\begin{equation}
\label{eq:mlp}
\mathrm{MLP}_n(\underset{s\times h}{\mathrm{X}}) =
\mathrm{ReLU}(
\underset{s\times h}{\mathrm{X}} \times \underset{h\times p}{\mathbf{W}^{l1}_n} + \underset{s\times p}{\mathbf{B}^{l1}_n}) \times \underset{p\times h}{\mathbf{W}^{l2}_n}
+ \underset{s\times h}{\mathbf{B}^{l2}_n},
\end{equation}

where
$\mathbf{W}^{l1}_n$ is the matrix of parameters of the first fully connected layer and $\mathbf{B}^{l1}_n$ are its bias parameters broadcasted along the sequence dimension: $\underset{s\times h}{\mathbf{B}^{l1}_n} = \underset{s\times 1}{\mathbf{1}} \times \underset{1\times h}{\mathbf{b}^{l1}_n}$.
$\mathbf{W}^{l2}_n$ and $\mathbf{B}^{l2}_n$ are the parameters of the second fully connected layer.
The broadcast operator applied to the bias parameters is omitted for simplicity.
The size of the internal dimension of the MLP component is represented with $p$.
The considered architecture instantiation assumes the uses of $\mathrm{ReLU}(\cdot)$ \citep{Glorot2011DeepSR} as a non-linearity function as this is a common choice. The proposed transformations also maintain the function preserving property with alternative choices such as $\mathrm{GELU}(\cdot)$ \citep{Hendrycks2016GaussianEL}.

$\mathrm{MHA}_n(\cdot)$ is the \emph{Multi Head Attention} defined as:

\begin{equation}
\label{eq:mha}
\begin{array}{l}
\mathrm{MHA}_n(\underset{s\times h}{\mathrm{X}}) = \left [ \underset{s\times v}{\mathrm{H}_1} \cdots \ \underset{s\times v}{\mathrm{H}_E} \right ] \times \underset{(E\cdot v) \times h}{\mathbf{W}^{O}_n},
\\
\\
\underset{s\times v}{\mathrm{H}_e} = \mathrm{Attention}(
\underset{s\times h}{\mathrm{X}}\!\!\times\!\underset{h\times k}{\mathbf{W}^{Q}_{n,e}},
\underset{s\times h}{\mathrm{X}}\!\!\times\!\underset{h\times k}{\mathbf{W}^{K}_{n,e}},
\underset{s\times h}{\mathrm{X}}\!\!\times\!\underset{h\times v}{\mathbf{W}^{V}_{n,e}}
)
\ \ \ \ \forall \ e\in [1, E],
\\
\\
\mathrm{Attention}(\underset{s\times k}{\mathrm{Q}},\underset{s \times k}{\mathrm{K}},\underset{s\times v}{\mathrm{V}}) = \mathrm{Softmax}( \frac{1}{\sqrt{k}} \cdot \underset{s \times k}{\mathrm{Q}}\times \underset{k \times s}{\mathrm{K}^\top} )  \times \underset{s \times v}{\mathrm{V}},
\end{array}
\end{equation}

where
$E$ is the number of heads,
$k$ is the hidden dimension of \emph{key}, $\mathrm{K}$, and \emph{query}, $\mathrm{Q}$,
and $v$ is the hidden dimension of \emph{value}, $\mathrm{V}$.
$\mathrm{K}^\top$ represents the transpose of $\mathrm{K}$. The concatenation of the representations produced by the attention heads is represented with the \emph{block notation}: $\mathrm{C} = \left [ \mathrm{A}  \ \ \mathrm{B} \right ] $.

As the normalization function in each component, we use RMSNorm \citep{Zhang2019RootMS}. The original definition of the transformer architecture uses LayerNorm, but RMSNorm has become a more common design choice in large language models \citep{Raffel2020ExploringTL, Rae2021ScalingLM, touvron2023llama}. The key difference is only scaling the variance of the inputs and using scaling parameters, rather than also subtracting their mean and using bias parameters.
Thus, we define $\mathrm{Norm}(\cdot)$ as:

\begin{equation} \label{eq:norm}
    \mathrm{Norm}_n^c(\underset{s \times h}{\mathrm{X}}) =
    % \underset{s \times h}{\mathrm{Y}} :=
    % \begin{bmatrix}
    \biggr [\frac{x_{i,j} \cdot \mathrm{g}_{n,j}^c}{\sqrt{\frac{1}{h} \sum_{\gamma = 1}^{h} (x_{i,\gamma})^2} } \ \ \ \vert \ i\!\in\![1, s] \land j\!\in
    \![1, h]
    \biggr ]
    \ \ \forall n\!\in\![1,N] \land c\!\in\!\{\mathrm{MHA},\mathrm{MLP}\},
    % + \varepsilon
    % \end{bmatrix}   
\end{equation}

where
$\underset{1\times h}{\mathbf{g}_n^c}$ identifies the vector of the scaling parameters of the $\mathrm{Norm}(\cdot)$ instance of component $c$ in the $n^\text{th}$ layer.
%$\varepsilon \simeq 0$ is a positive value added to avoid division by 0.

\section{Function preserving transformations} % of the Transformer architecture}
\label{section:method}

In this section, we define six \emph{function preserving transformations} that can be applied to extend a transformer architecture to increase its scale while keeping its function unaltered, thus allowing to introduce new parameters to store additional knowledge while preserving the knowledge acquired so far. Each transformation is defined to target the expansion of one of the hyper-parameters of the architecture: $p, E, v, k, h,$ and $N$, each controlling a distinct dimension of the scaling. The proposed transformations are summarized in Table \ref{table:summary1}. 

For each transformation, we define how the existing parameters must be expanded and propose a set of minimal initialization constraints to obtain the function preserving property with proof.

The presented transformations can be combined to allow the joint extension of multiple dimensions of the transformer architecture. Furthermore, different subsets of such transformations can be applied incrementally, interleaving training iterations, as well as independently to different parts of the architecture.

Symbols denoting parameters, representations, and functions resulting from the application of the transformation discussed in each of the following subsection are indicated with the ``hat'' symbol: \^{}.

\begin{table}[]\small
\hyphenpenalty=500
\renewcommand{\arraystretch}{1.2}
\makebox[\textwidth]{
\begin{tabular}{|p{1.25cm}|p{8.9cm}|p{5.9cm}|}
\hline
Name & Transformation & Function preserving constraint \\ \hline 
Sec. \ref{ss:mlpexpansion}: MLP\newline expansion & Def. \ref{def:mlpexpansion}: to increase the MLP internal dimension $p$ to $\hat{p},$ add $\hat{p}-p$ columns to the the first MLP weight matrix and bias vector and add $\hat{p}-p$ rows to the second MLP weight matrix.
 & Thrm. \ref{thm:mlpexpansion}: zero initialize the new $\hat{p}-p$ rows of the second MLP weight matrix.
\\ \hline
Sec. \ref{ss:headaddition}:   Head\newline addition  & Def. \ref{def:headaddition}: to increase the number of attention heads $E$, per head added, add $v$ rows to the MHA output weight matrix.
 & Thrm. \ref{thm:headaddition}: zero initialize the new $v$ rows of the MHA output weight matrix.
 \\ \hline
Sec. \ref{ss:headsexpansion}:   Heads\newline expansion  &  Def. \ref{def:headsexpansion}: to increase the attention head representation dimension $v$ to $\hat{v},$ add $\hat{v}-v$ columns to the value weight matrix and insert $\hat{v}-v$ rows to each of $E$ splits of the MHA output weight matrix.
& Thrm. \ref{thm:headsexpansion}: zero initialize the new $\hat{v}-v$ rows inserted to each of $E$ splits of the MHA output weight matrix.
\\ \hline
Sec. \ref{ss:attnexpansion}:   Attention\newline expansion  & Def. \ref{def:attnexpansion}: to increase the key/query representation dimension $k$ to $\hat{k},$ add $\hat{k}-k$ columns to the key/query weight matrices and scale the key weight matrix by ${\sqrt{\hat{k}}}/{\sqrt{k}}.$
& Thrm. \ref{thm:attnexpansion}: zero initialize the new $\hat{k}-k$\rule{0pt}{2.6ex} columns of the key weight matrix. 
\\ \hline
Sec. \ref{ss:hiddendimexpansion}:  Hidden\newline dimension\newline expansion  & Def. \ref{def:hiddendimexpansion}: to increase the transformer hidden dimension $h$ to $\hat{h},$ add $\hat{h}-h$ columns to the positional encoding matrix, norm scaling vector, second MLP weight matrix and bias vector, MHA output weight matrix, and input representation matrix; add $\hat{h}-h$ rows to the transformer output weight matrix, first MLP weight matrix, and key/query/value weight matrices; scale norm scaling vector by ${\sqrt{h}}/{\sqrt{\hat{h}}}.$ & Thrm. \ref{thm:hiddendimexpansion}: zero initialize the new $\hat{h}-h$\rule{0pt}{2.6ex}  columns of the positional encoding matrix, norm scaling vector, second MLP weight matrix and bias vector, and MHA output weight matrix.  \\ \hline
Sec. \ref{ss:layeraddition}: Layer\newline addition & Def. \ref{def:layeraddition}: to increase the number of layers $N$ to $\hat N,$ per layer added, insert new layer at position $n$ and increment index of all following layers.
&  Thrm. \ref{thm:layeraddition}: zero initialize the new layer's MHA output weight matrix and weight matrix and bias vector of the second MLP layer.
 \\ \hline
\end{tabular}}
\caption{Summary of proposed function preserving transformations.}
\label{table:summary1}
\end{table}

\subsection{MLP expansion} \label{ss:mlpexpansion}

The \emph{MLP expansion} transformation can be applied to expand the scale of the MLP by expanding the dimension of its internal representation.
This scaling dimension is controlled by the hyper-parameter $p$ introduced in Equation \ref{eq:mlp}.

\begin{definition}[MLP expansion] \label{def:mlpexpansion}
Given a Transformer model as defined in Section \ref{sec:trns},
the internal dimension of $\mathrm{MLP}_n \ \forall \ n \! \! \in \! \! [1, N]$ can be increased from $p$ to $\hat{p}$ by applying the following parameter-matrix transformations:

% \begin{equation}
% \begin{array}{l}
\begin{align*}
    &\underset{h\times p}{\mathbf{W}^{l1}_n} \mapsto  \underset{h\times \hat{p}}{\mathbf{\hat{W}}^{l1}_n}
    := \left [ \underset{h\times p}{\mathbf{W}^{l1}_n} \ \ \ \underset{h\times (\hat{p}-p)}{\mathbf{M}^{Wl1}_n}\right ],\stepcounter{equation}\tag{\theequation}
\\
\\
    &\underset{1\times p}{\mathbf{b}^{l1}_n} \mapsto \underset{1\times \hat{p}}{\mathbf{\hat{b}}^{l1}_n}:= \left [ \underset{1\times p}{\mathbf{b}^{l1}_n} \ \ \ \underset{1\times (\hat{p}-p)}{\mathbf{m}^{bl1}_n}\right ],\stepcounter{equation}\tag{\theequation}
\\
\\
    &\underset{p\times h}{\mathbf{W}^{l2}_n} \mapsto
    \underset{\hat{p}\times h}{\mathbf{\hat{W}}^{l2}_n}
    := \left [ 
    \begin{array}{c}
    \underset{p\times h}{\mathbf{W}^{l2}_n}
    \\
    \\
    \underset{(\hat{p}-p)\times h}{\mathbf{M}^{Wl2}_n}
    \end{array}
    \right ],\stepcounter{equation}\tag{\theequation}
\end{align*}
% \end{array}
% \end{equation}

where $\underset{h\times (\hat{p}-p)}{\mathbf{M}^{Wl1}_n},$ $\underset{1\times (\hat{p}-p)}{\mathbf{m}^{bl1}_n},$ and $\underset{(\hat{p}-p)\times h}{\mathbf{M}^{Wl2}_n}$ are matrices of the specified shape. 
For the purpose of defining of the MLP expansion transformation, the values of these matrices can be assumed to be arbitrary. Constraints on their \emph{initializer functions} are introduced below to achieve the function preserving property.

No other modifications to the Transformer architecture are required since the $\mathrm{MLP}_n(\cdot)$ function (Equation \ref{eq:mlp}) still inputs and outputs matrices of shape $s \times h$ after the transformation.

\begin{flushright}
\qedsymbol
\end{flushright}
\end{definition}

\begin{theorem}[Function preserving MLP expansion] \label{thm:mlpexpansion}

\begin{align*}
&\underset{(\hat{p}-p)\times h}{\mathbf{M}^{Wl2}_n} := \underset{(\hat{p}-p)\times h}{\mathbf{0}} \stepcounter{equation}\tag{\theequation}
\end{align*}

$\implies$

\begin{align*}
\mathrm{ReLU}(
\underset{s\times h}{\mathrm{X}} \times \underset{h\times p}{\mathbf{W}^{l1}_n} + \underset{s\times p}{\mathbf{B}^{l1}_n}) \times \underset{p\times h}{\mathbf{W}^{l2}_n}
+ \underset{s\times h}{\mathbf{B}^{l2}_n} = \mathrm{ReLU}(
\underset{s\times h}{\mathrm{X}} \times \underset{h\times p}{\mathbf{\hat{W}}^{l1}_n} + \underset{s\times p}{\mathbf{\hat{B}}^{l1}_n}) \times \underset{p\times h}{\mathbf{\hat{W}}^{l2}_n}
+ \underset{s\times h}{\mathbf{B}^{l2}_n}\stepcounter{equation}\tag{\theequation}
\end{align*}

Informally:
zero initializing $\underset{(\hat{p}-p)\times h}{\mathbf{M}^{Wl2}_n}$
implies the \emph{function preservation} property for the MLP expansion transformation.
\end{theorem}

See Appendix \ref{app:mlpexpansion} for proof.

The MLP expansion transformation can be applied to all the MLP blocks to maintain the MLP internal dimension uniformly across all the layers. However, it can also be applied to only a subset of the layers independently to allow experimenting with different capacity at different depths.

\subsection{Head addition}\label{ss:headaddition}

The \emph{Head addition} transformation can be applied to add new heads in a MHA component.
This scaling dimension is controlled by the hyper-parameter $E$ introduced in Equation \ref{eq:mha}.

\begin{definition}[Head addition] \label{def:headaddition}
Given a Transformer model as defined in Section \ref{sec:trns},
a new head can be added to $\mathrm{MHA}_n(\cdot) \  \forall \ n \! \! \in \! \! [1, N]$ by introducing new input projection matrices: $\underset{h\times k}{\mathbf{W}^{Q}_{n,E+1}},
\underset{h\times k}{\mathbf{W}^{K}_{n,E+1}}, \underset{h\times v}{\mathbf{W}^{V}_{n,E+1}}$ and applying the following parameter-matrix transformation to the output projection matrix:

\begin{equation}
    \underset{(E\cdot v) \times h}{\mathbf{W}^{O}_n} \mapsto
    \underset{((E+1)\cdot v) \times h}{\mathbf{\hat{W}}^{O}_n}
    := \left [ 
    \begin{array}{c}
    \underset{(E\cdot v) \times h}{\mathbf{W}^{O}_n}
    \\
    \\
    \underset{v\times h}{\mathbf{M}^{W\!O}_n}
    \end{array}
    \right ].
\end{equation}

No other modifications to the Transformer architecture are required since the $\mathrm{MHA}_n(\cdot)$ function (Equation \ref{eq:mha}) still inputs and outputs matrices of shape $s \times h$ after the transformation.

\begin{flushright}
\qedsymbol
\end{flushright}
\end{definition}

The \emph{Head addition} transformation is defined to add one new head.
The transformation can be applied multiple times to add an arbitrary number of new heads.

\begin{theorem}[Function preserving head addition] \label{thm:headaddition}

\begin{equation}
\underset{v\times h}{\mathbf{M}^{W\!O}_n}
 := \underset{v\times h}{\mathbf{0}}
\implies
\left [ \underset{s\times v}{\mathrm{H}_1} \cdots \ \underset{s\times v}{\mathrm{H}_E} \right ] \times \underset{(E\cdot v) \times h}{\mathbf{W}^{O}_n}
=
\left [ \underset{s\times v}{\mathrm{H}_1} \cdots \ \underset{s\times v}{\mathrm{H}_{(E+1)}} \right ] \times \underset{((E+1)\cdot v) \times h}{\mathbf{\hat{W}}^{O}_n}
\end{equation}

Informally:
zero initializing $\underset{v\times h}{\mathbf{M}^{W\!O}_n}$
implies the \emph{function preservation} property for the head addition transformation.
\end{theorem}

See Appendix \ref{app:headaddition} for proof.

The head addition transformation can be applied to all the MHA blocks to maintain the number of MHA heads uniformly across all the layers. However, it can also be applied to only a subset of the layers independently to allow experimenting with different capacity at different depths.

\subsection{Heads expansion} \label{ss:headsexpansion}

The \emph{Heads expansion} transformation can be applied to expand the dimension of the representation generated by each attention heads.
This scaling dimension is controlled by the hyper-parameter $v$ introduced in Equation \ref{eq:mha}.

\begin{definition}[Heads expansion] \label{def:headsexpansion}
Given a Transformer model as defined in Section \ref{sec:trns},
the dimension of representation generated by the attention heads, $\underset{s\times v}{\mathrm{H}_e} \ \forall \ e \! \! \in \! \! [1, E]$, of $\mathrm{MHA}_n \ \forall \ n \! \! \in \! \! [1, N]$ can be increased from $v$ to $\hat{v}$ by applying the following parameter-matrix transformations:

\begin{align*}
&\underset{h\times v}{\mathbf{W}^{V}_{n,e}}
\mapsto
\underset{h\times \hat{v}}{\mathbf{\hat{W}}^{V}_{n,e}}
:=
\left [
\underset{h\times v}{\mathbf{W}^{V}_{n,e}}
\ \ \ 
\underset{h\times (\hat{v} - v)}{\mathbf{M}^{W\!V}_{n,e}}
\right ]
\ \ \ \forall \ e\in [1, E], \stepcounter{equation}\tag{\theequation}
\\
\\
&\underset{v \times h}{\mathbf{W}^{O}_{n,e}}
\mapsto
\underset{\hat{v} \times h}{\mathbf{\hat{W}}^{O}_{n,e}}
:=
\left [ 
    \begin{array}{c}
    \underset{v \times h}{\mathbf{W}^{O}_{n,e}}
    \\
    \\
    \underset{(\hat{v}-v)\times h}{\mathbf{M}^{W\!O}_{n,e}}
    \end{array}
\right ]
\ \ \ \forall \ e\in [1, E], \stepcounter{equation}\tag{\theequation}
\end{align*}

where $\underset{v \times h}{\mathbf{W}^{O}_{n,e}}$ is the $e^\text{th}$ ``split'' of $\underset{(E \cdot v) \times h}{\mathbf{W}^{O}_{n}}$ along the $(E\cdot v)$ dimension:

\begin{equation}
\underset{(E \cdot v) \times h}{\mathbf{W}^{O}_{n}}
:=
\left [
    \begin{array}{c}
    \vdots
    \\
    \underset{v \times h}{\mathbf{W}^{O}_{n,e}}
    
    \\
    \vdots
    \end{array}
    \begin{array}{r}
     \ \vert \ \ e\in [1, E].
    %  \ \forall \ \ e\in [1, E]
     \end{array}
\right ]
\end{equation}

No other modifications to the Transformer architecture are required since the $\mathrm{MHA}_n(\cdot)$ function (Equation \ref{eq:mha}) still inputs and outputs matrices of shape $s \times h$ after the transformation.

\begin{flushright}
\qedsymbol
\end{flushright}
\end{definition}

\begin{theorem}[Function preserving heads expansion] \label{thm:headsexpansion}

\begin{equation}
\underset{(\hat{v}-v)\times h}{\mathbf{M}^{W\!O}_{n,e}}
:= 
\underset{(\hat{v}-v)\times h}{\mathbf{0}}
\implies
\left [ \underset{s\times v}{\mathrm{H}_1} \cdots \ \underset{s\times v}{\mathrm{H}_E} \right ] \times \underset{(E\cdot v) \times h}{\mathbf{W}^{O}_n}
=
\left [ \underset{s\times \hat{v}}{\mathrm{\hat{H}}_1} \cdots \ \underset{s\times \hat{v}}{\mathrm{\hat{H}}_{E}} \right ] \times \underset{(E\cdot \hat{v}) \times h}{\mathbf{\hat{W}}^{O}_n}
\end{equation}

where:

\begin{equation}
    \underset{s\times \hat{v}}{\mathrm{\hat{H}}_e} = \mathrm{Attention}(
\underset{s\times h}{\mathrm{X}}\!\!\times\!\underset{h\times k}{\mathbf{W}^{Q}_{n,e}},
\underset{s\times h}{\mathrm{X}}\!\!\times\!\underset{h\times k}{\mathbf{W}^{K}_{n,e}},
\underset{s\times h}{\mathrm{X}}\!\!\times\!\underset{h\times \hat{v}}{\mathbf{\hat{W}}^{V}_{n,e}}
)
\end{equation}

Informally:
zero initializing $\underset{(\hat{v}-v)\times h}{\mathbf{M}^{W\!O}_{n,e}}$
implies the \emph{function preservation} property for the head expansion transformation.
\end{theorem}

See Appendix \ref{app:headsexpansion} for proof

The heads expansion transformation can be applied to all heads of all the MHA blocks to maintain the attention head representation dimension uniformly across all the layers. However, it can also be applied to only a subset of the layers or even a subset of attention heads independently to allow experimenting with different capacity at different parts of the architecture.

\subsection{Attention expansion} \label{ss:attnexpansion}
The \emph{Attention expansion} transformation can be applied to expand the \emph{key} and \emph{query} representations whose inner product produces the attention weights matrix.
This scaling dimension is controlled by the hyper-parameter $k$ introduced in Equation \ref{eq:mha}.

\begin{definition}[Attention expansion] \label{def:attnexpansion}
Given a Transformer model as defined in Section \ref{sec:trns},
the dimension of representations generating the attention weights of $\mathrm{MHA}_n \ \forall \ n \!  \in  \! [1, N]$ can be increased from $k$ to $\hat{k}$ by applying the following parameter-matrix transformations:

%\begin{equation}
%\begin{array}{l}
\begin{align*}
&\underset{h\times k}{\mathbf{W}^{Q}_{n,e}}
\mapsto
\underset{h\times \hat{k}}{\mathbf{\hat{W}}^{Q}_{n,e}}
:=
\left [
\underset{h\times k}{\mathbf{W}^{Q}_{n,e}}
\ \ \ 
\underset{h\times (\hat{k} - k)}{\mathbf{M}^{W\!Q}_{n,e}}
\right ]
\ \ \ \forall \ e\in [1, E],\stepcounter{equation}\tag{\theequation}
\\
\\
&\underset{h\times k}{\mathbf{W}^{K}_{n,e}}
\mapsto
\underset{h\times \hat{k}}{\mathbf{\hat{W}}^{K}_{n,e}}
:=
\left [
 \frac{\sqrt{\hat{k}}}{\sqrt{k}} %\frac{\sqrt{\hat{k}}}{\sqrt{k}} or \sqrt{\frac{\hat{k}}{k}}
 \cdot 
 \underset{h\times k}{\mathbf{W}^{K}_{n,e}}
\ \ \ 
\underset{h\times (\hat{k} - k)}{\mathbf{M}^{W\!K}_{n,e}}
\right ]
\ \ \ \forall \ e\in [1, E].\stepcounter{equation}\tag{\theequation}\label{eqn:scalek}
\end{align*}
%\end{array}
%\end{equation}

\begin{flushright}
\qedsymbol
\end{flushright}
\end{definition}

\begin{theorem}[Function preserving attention expansion] \label{thm:attnexpansion}

\begin{align*}
\underset{h\times (\hat{k}-k)}{\mathbf{M}^{W\!K}_{n,e}}
:= 
\underset{h\times (\hat{k}-k)}{\mathbf{0}} \stepcounter{equation}\tag{\theequation}
\end{align*}

$\implies$

\begin{align*}
\mathrm{Attention}(
\underset{s\times h}{\mathrm{X}}\!\!\times\!\underset{h\times k}{\mathbf{W}^{Q}_{n,e}},
\underset{s\times h}{\mathrm{X}}\!\!\times\!\underset{h\times k}{\mathbf{W}^{K}_{n,e}},
\underset{s\times h}{\mathrm{X}}\!\!\times\!\underset{h\times v}{\mathbf{W}^{V}_{n,e}}
) 
=
\mathrm{Attention}(
\underset{s\times h}{\mathrm{X}}\!\!\times\!\underset{h\times \hat{k}}{\mathbf{\hat{W}}^{Q}_{n,e}},
\underset{s\times h}{\mathrm{X}}\!\!\times\!\underset{h\times \hat{k}}{\mathbf{\hat{W}}^{K}_{n,e}},
\underset{s\times h}{\mathrm{X}}\!\!\times\!\underset{h\times v}{\mathbf{W}^{V}_{n,e}}
)\stepcounter{equation}\tag{\theequation}
\end{align*}

Informally:
zero initializing $\underset{h\times (\hat{k}-k)}{\mathbf{M}^{W\!K}_{n,e}}$
implies the \emph{function preservation} property for the attention expansion transformation.
\end{theorem}

See Appendix \ref{app:attnexpansion} for proof.

In most transformer implementations, $k = v$. In such cases, the attention expansion may be performed jointly with the head expansion.

The attention expansion transformation can be applied to all heads of all the MHA blocks to maintain the key/query representation dimension uniformly across all the layers. However, it can also be applied to only a subset of the layers or even a subset of attention heads independently to allow experimenting with different capacity at different parts of the architecture.

\subsection{Hidden dimension expansion} \label{ss:hiddendimexpansion}

The \emph{Hidden dimension expansion} transformation can be applied to expand the dimension of the representation produced by the transformer layers.
This scaling dimension is controlled by the hyper-parameter $h$ introduced in Equation \ref{eq:layers}.

\begin{definition}[Hidden dimension expansion] \label{def:hiddendimexpansion}
Given a Transformer model as defined in Section \ref{sec:trns},
the dimension of the transformer layers' input/output representation can be increased from $h$ to $\hat{h}$ by applying the following parameter-matrix transformations:
\begin{align*}
& \underset{s\times h}{\mathbf{P}}
\mapsto
\underset{s\times \hat{h}}{\mathbf{\hat{P}}}
:=
\left [
\underset{s\times h}{\mathbf{P}}
\ \ \ 
\underset{s\times (\hat{h} - h)}{\mathbf{M}^{P}}
\right ],\stepcounter{equation}\tag{\theequation}
\\
\\
& \underset{h\times o}{\mathbf{W}^{out}}
\mapsto
\underset{\hat{h}\times o}{\mathbf{\hat{W}}^{out}}
:=
\left [
\begin{array}{c}
\underset{h\times o}{\mathbf{W}^{out}}
  \\
  \\
\underset{(\hat{h} - h)\times o}{\mathbf{M}^{Wout}}
\end{array}
\right ],\stepcounter{equation}\tag{\theequation}
\\
\\
& \underset{1\times h}{\mathbf{g}_n^c}
\mapsto
\underset{1\times \hat{h}}{\mathbf{\hat{g}}_n^c}
:=
\left [
\frac{\sqrt{h}}{\sqrt{\hat{h}}}
\cdot
\underset{1\times h}{\mathbf{g}_n^c}
\ \ \
\underset{1 \times (\hat{h} - h)}{\mathbf{m}^{g,c}_n}
\right ]
    \ \ \forall n\!\in\![1,N] \land c\!\in\!\{\mathrm{MHA},\mathrm{MLP}\},\stepcounter{equation}\tag{\theequation}\label{eqn:scaleh}
\\
\\
& \underset{h\times p}{\mathbf{W}^{l1}_n}
\mapsto
\underset{\hat{h}\times p}{\mathbf{\hat{W}}^{l1}_n}
:=
\left [
\begin{array}{c}
\underset{h\times p}{\mathbf{W}^{l1}_n}
  \\
  \\
\underset{(\hat{h} - h)\times p}{\mathbf{M}^{Wl1}}
\end{array}
\right ]
    \ \ \forall n\!\in\![1,N],\stepcounter{equation}\tag{\theequation}
\\
\\
& \underset{p\times h}{\mathbf{W}^{l2}_n}
\mapsto
\underset{p\times \hat{h}}{\mathbf{\hat{W}}^{l2}_n}
:=
\left [
\underset{p\times h}{\mathbf{W}^{l2}_n}
\ \ \
\underset{p \times (\hat{h} - h)}{\mathbf{M}^{Wl2}_{n}}
\right ]
    \ \ \forall n\!\in\![1,N],\stepcounter{equation}\tag{\theequation}
\\
\\
& \underset{1\times h}{\mathbf{b}^{l2}_n}
\mapsto
\underset{1\times \hat{h}}{\mathbf{\hat{b}}^{l2}_n}
:=
\left [
\underset{1\times h}{\mathbf{b}^{l2}_n}
\ \ \
\underset{1 \times (\hat{h} - h)}{\mathbf{m}^{bl2}_{n}}
\right ]
    \ \ \forall n\!\in\![1,N],\stepcounter{equation}\tag{\theequation}
\\
\\
& \underset{h\times k}{\mathbf{W}^{Q}_{n,e}}
\mapsto
\underset{\hat{h}\times k}{\mathbf{\hat{W}}^{Q}_{n,e}}
:=
\left [
\begin{array}{c}
\underset{h\times k}{\mathbf{W}^{Q}_{n,e}}
  \\
  \\
\underset{(\hat{h} - h)\times k}{\mathbf{M}^{W\!Q}_{n,e}}
\end{array}
\right ]
    \ \ \forall n\!\in\![1,N] \land e\!\in\![1,E],\stepcounter{equation}\tag{\theequation}
\\
\\
& \underset{h\times k}{\mathbf{W}^{K}_{n,e}}
\mapsto
\underset{\hat{h}\times k}{\mathbf{\hat{W}}^{K}_{n,e}}
:=
\left [
\begin{array}{c}
\underset{h\times k}{\mathbf{W}^{K}_{n,e}}
  \\
  \\
\underset{(\hat{h} - h)\times k}{\mathbf{M}^{W\!K}_{n,e}}
\end{array}
\right ]
    \ \ \forall n\!\in\![1,N] \land e\!\in\![1,E],\stepcounter{equation}\tag{\theequation}
\\
\\
& \underset{h\times v}{\mathbf{W}^{V}_{n,e}}
\mapsto
\underset{\hat{h}\times v}{\mathbf{\hat{W}}^{V}_{n,e}}
:=
\left [
\begin{array}{c}
\underset{h\times v}{\mathbf{W}^{V}_{n,e}}
  \\
  \\
\underset{(\hat{h} - h)\times v}{\mathbf{M}^{W\!V}_{n,e}}
\end{array}
\right ]
    \ \ \forall n\!\in\![1,N] \land e\!\in\![1,E],\stepcounter{equation}\tag{\theequation}
\\
\\
& \underset{(E\cdot v) \times h}{\mathbf{W}^{O}_n}
\mapsto
\underset{(E\cdot v) \times \hat{h}}{\mathbf{\hat{W}}^{O}_n}
:=
\left [
\underset{(E\cdot v) \times h}{\mathbf{W}^{O}_n}
\ \ \ 
\underset{(E\cdot v)\times (\hat{h} - h)}{\mathbf{M}^{W\!O}_{n}}
\right ]
    \ \ \forall n\!\in\![1,N],\stepcounter{equation}\tag{\theequation}
\end{align*}

and modifying the embedding function to produce an extended input representation:
\begin{equation}
\underset{s\times \hat{h}}{\mathrm{\hat{I}}}
:=
\left [
\underset{s\times h}{\mathrm{I}}
\ \ \ 
\underset{s\times (\hat{h} - h)}{\mathrm{M}^I}
\right ].
\end{equation}
For example, a token embedding table can be expanded by adding $(\hat{h} -h)$ randomly initialized columns, mapping the same vocabulary into an extended embedding.
% TODO add example for image input embedding?

\begin{flushright}
\qedsymbol
\end{flushright}
\end{definition}

\begin{theorem}[Function preserving hidden dimension expansion] \label{thm:hiddendimexpansion}

\begin{align*}
&\underset{s\times (\hat{h} - h)}{\mathbf{M}^{P}}
:= 
\underset{s\times (\hat{h} - h)}{\mathbf{0}}\stepcounter{equation}\tag{\theequation}
\\
\\
&\underset{p \times (\hat{h} - h)}{\mathbf{M}^{Wl2}_{n}}
:= 
\underset{p \times (\hat{h} - h)}{\mathbf{0}}
    \ \ \forall n\!\in\![1,N]\stepcounter{equation}\tag{\theequation}
\\
\\
&\underset{1 \times (\hat{h} - h)}{\mathbf{m}^{bl2}_{n}}
:= 
\underset{1 \times (\hat{h} - h)}{\mathbf{0}}
    \ \ \forall n\!\in\![1,N]\stepcounter{equation}\tag{\theequation}
\\
\\
&\underset{(E\cdot v)\times (\hat{h} - h)}{\mathbf{M}^{W\!O}_{n}}
:= 
\underset{(E\cdot v) \times (\hat{h} - h)}{\mathbf{0}}
    \ \ \forall n\!\in\![1,N]\stepcounter{equation}\tag{\theequation}
\\
\\
&\underset{s\times (\hat{h} - h)}{\mathrm{M}^I}
:=
\underset{s\times (\hat{h} - h)}{0}\stepcounter{equation}\tag{\theequation}
\end{align*}

$\implies$

\begin{align*}
&\underset{s\times \hat{h}}{\mathrm{\hat{I}}_n} = [\underset{s\times h}{\mathrm{I}_n} \ \ \ \underset{s\times (\hat{h}-h)}{\mathrm{0}}]
\ \ \ \ \ \forall n\!\in\![1,N+1]\stepcounter{equation}\tag{\theequation}
\end{align*}

$\implies$

\begin{align*}
&\mathrm{TransformerLayer}^{\circ N}(\underset{s\times h}{\mathrm{I}}\!+\!\underset{s\times h}{\mathbf{P}})\ \times \underset{h\times o}{\mathbf{W}^{out}}
=
\mathrm{\hat{TransformerLayer}}^{\circ N}(\underset{s\times h}{\mathrm{I}}\!+\!\underset{s\times \hat{h}}{\mathbf{\hat{P}}})\ \times \underset{\hat{h}\times o}{\mathbf{\hat{W}}^{out}}\stepcounter{equation}\tag{\theequation}
\end{align*}

where
$\underset{s\times h}{\mathrm{I}_{N+1}}$ refers to the representations outputted by the last transformer layer,
and $\underset{s\times h}{\mathrm{I}_n}\ \forall n\!\in\![1,N]$ refers to the representation inputted by the $n^{th}$ transformer layer.
Symbols denoting parameters, representations and functions resulting from the application of the transformation discussed in this section are indicated with the ``hat'' \^{} symbol.

Informally: 
zero initializing the specified matrices
implies the \emph{function preservation} property for the hidden dimension expansion transformation.
\end{theorem}

See Appendix \ref{app:hiddendimexpansion} for proof.

The hidden dimension expansion transformation must be applied to all MHA blocks to maintain the hidden dimension uniformly across all the layers, due to the skip connections used throughout the architecture.

\subsection{Layer addition} \label{ss:layeraddition}
The \emph{Layer addition} transformation can be applied to insert an new layer at any depth of the current Transformer architecture.
This scaling dimension is controlled by the hyper-parameter $N$ introduced in Equation \ref{eq:layers}.

\begin{definition}[Layer addition] \label{def:layeraddition}
A new $\mathrm{TransformerLayer}(\cdot)$ whose parameters allow to input and output matrices of $x \times h$ can be inserted in the sequence of the pre-existing $N$ layers.
The new transformer layer can be inserted at any position $n \in [1, N\!+\!1]$. The index of the downstream layers is incremented by one.
\begin{flushright}
\qedsymbol
\end{flushright}
\end{definition}

\begin{theorem}[Function preserving layer addition] \label{thm:layeraddition}

With $n$ being the index of the added layer:
\begin{equation} \left. 
\begin{aligned} 
\underset{(E\cdot v) \times h}{\mathbf{W}^{O}_n} &:= \underset{(E\cdot v) \times h}{\mathbf{0}}
\\
\hfill 
\underset{p\times h}{\mathbf{W}^{l2}_n} &:= \underset{p\times h}{\mathbf{0}}
\\
\hfill 
\underset{1\times h}{\mathbf{b}^{l2}_n} &:= \underset{1\times h}{\mathbf{0}}
\end{aligned} \right\}
\implies \mathrm{TransformerLayer}_n(\underset{s\times h}{\mathrm{I}_n}) 
= \underset{s\times h}{\mathrm{I}_n}
\end{equation}

Informally:
Zero initializing the parameters of the output projections of the MLP and MHA implies that the added transformer layer output is equivalent to the input.
\end{theorem}

See Appendix \ref{app:layeraddition} for proof.

\section{Related work}
\label{section:related}
Some existing works have proposed function preserving transformer expansion operators, but none cover all six dimensions as proposed in this work. Bert2BERT \citep{chen2022bert2bert} proposes function preserving width expansions of the MLP internal dimension, hidden dimension, and number of attention heads. \citet{shen2022staged} achieve function preserving width expansion, although constrained to doubling of all matrix and vector dimensions, and depth expansion via zero initialization of LayerNorm and bias parameters. \citet{yao20232x} use masking on new hidden MLP neurons, attention heads, and layers to achieve function preservation. \citet{wang2023learning} use an inner optimization to learn a linear mapping for parameter expansion in depth and width, but without constraints for function preservation. Notably, our transformations form a function preserving subspace of their learnable space.  Deep Fusion \citep{mazzawi2023deep} extends the concept of expansion to multiple source models, where the special case of self-fusion achieves function preserving width expansion. Of these works, some methods are nearly function preserving but admit gaps due to LayerNorm discrepancies \citep{chen2022bert2bert, mazzawi2023deep}. No known works consider scaling factors, as we address in Equations \ref{eqn:scalek} and \ref{eqn:scaleh}, nor RMSNorm.

\section{Conclusion}

We have defined six transformations that can be applied to a transformer model to increase the scale of all the different aspects of the architecture:
1) size of MLP internal representation,
2) number of attention heads,
3) size of the attention heads output representation,
4) size of the attention input representation,
5) size of the transformer layers input/output representations, 
6) number of layers.
For each of these transformations, we have provided a proof of exact function preservation given a minimal set of constraints on the initialization of the added parameters. These six transformations are composable to permit many different ways to scale a transformer-based model while preserving its function.

We note that, there exist alternative definitions to such transformations that achieve function-preservation without requiring zero initialization.
However, the form of the proposed transformations is intended to be simple yet minimally constraining.
The space of possible initialization strategies may be explored with the aim to optimize for training in an empirical context.

In future work, these transformations may be applied in the training of a new large model by initializing a smaller model, training it under reduced data and computational complexity requirements, and incrementally scaling it to larger sizes throughout training to the desired final size. They may also be used to generate a family of models that are trained for the same task but at different sizes: all models within the family can begin from the same checkpoint from training the smallest model, then each successively sized model can be branched and finetuned at its final size. Finally, neural architecture search (NAS) techniques could be applied to determine optimal transformation scheduling and architectural progression for a given task and compute budget.    

\section{Acknowledgements}
We would like to thank Jeffrey Pennington and Utku Evci for their input to this work.

\vspace{8pt}
\bibliography{iclr2023_conference}
\bibliographystyle{iclr2023_conference}

%%%%%%%%%%%%%%%%%%%%%%%%%%%%%%%%%%%%%%%%%%%%%%%%%%%%%%%%%%%%

\clearpage
\appendix

\section{Proofs}

\subsection{MLP expansion} \label{app:mlpexpansion}

\begin{proof}
\begin{align*}
&
\mathrm{ReLU} (
\underset{s\times h}{\mathrm{X}} \times \underset{h\times p}{\mathbf{\hat{W}}^{l1}_n} + \underset{s\times p}{\mathbf{\hat{B}}^{l1}_n}) \times \underset{p\times h}{\mathbf{\hat{W}}^{l2}_n}
\\
&=
\mathrm{ReLU} \left(
\underset{s\times h}{\mathrm{X}} \times \left [ \underset{h\times p}{\mathbf{W}^{l1}_n}  \ \ \ \underset{h\times (\hat{p}-p)}{\mathbf{M}^{Wl1}_n} \right ]
+ \left [ \underset{1\times p}{\mathbf{B}^{l1}_n} \ \ \ \underset{1\times (\hat{p}-p)}{\mathbf{M}^{bl1}_n}\right ]\right) \times \left [ 
    \begin{array}{c}
    \underset{p\times h}{\mathbf{W}^{l2}_n}
    \\
    \\
    \underset{(\hat{p}-p)\times h}{\mathbf{0}}
    \end{array}
    \right ] \\
    \\
&=
\mathrm{ReLU} \left(
\left [ \underset{s\times h}{\mathrm{X}} \times  \underset{h\times p}{\mathbf{W}^{l1}_n}  \ \ \ \ \ \ \underset{s\times h}{\mathrm{X}} \times \underset{h\times (\hat{p}-p)}{\mathbf{M}^{Wl1}_n} \right ]
+ \left [ \underset{1\times p}{\mathbf{B}^{l1}_n} \ \ \ \underset{1\times (\hat{p}-p)}{\mathbf{M}^{bl1}_n}\right ]\right) \times \left [ 
    \begin{array}{c}
    \underset{p\times h}{\mathbf{W}^{l2}_n}
    \\
    \\
    \underset{(\hat{p}-p)\times h}{\mathbf{0}}
    \end{array}
    \right ] \\
\\
&= \mathrm{ReLU} \left(
\left [ \underset{s\times h}{\mathrm{X}} \times  \underset{h\times p}{\mathbf{W}^{l1}_n} + \underset{1\times p}{\mathbf{B}^{l1}_n} \ \ \ \ \ \ \underset{s\times h}{\mathrm{X}} \times \underset{h\times (\hat{p}-p)}{\mathbf{M}^{Wl1}_n} +\underset{1\times (\hat{p}-p)}{\mathbf{M}^{bl1}_n}
\right ]
\right) \times \left [ 
    \begin{array}{c}
    \underset{p\times h}{\mathbf{W}^{l2}_n}
    \\
    \\
    \underset{(\hat{p}-p)\times h}{\mathbf{0}}
    \end{array}
    \right ]
\\
\\
&= 
\left [ \mathrm{ReLU} (
\underset{s\times h}{\mathrm{X}} \times  \underset{h\times p}{\mathbf{W}^{l1}_n} + \underset{1\times p}{\mathbf{B}^{l1}_n}
) 
\ \ \ \ \ \ 
\mathrm{ReLU} (\underset{s\times h}{\mathrm{X}} \times \underset{h\times (\hat{p}-p)}{\mathbf{M}^{Wl1}_n} +\underset{1\times (\hat{p}-p)}{\mathbf{M}^{bl1}_n}
) 
\right ]
\times \left [ 
    \begin{array}{c}
    \underset{p\times h}{\mathbf{W}^{l2}_n}
    \\
    \\
    \underset{(\hat{p}-p)\times h}{\mathbf{0}}
    \end{array}
    \right ] \\
    \\
&= 
\left( \mathrm{ReLU} (
\underset{s\times h}{\mathrm{X}} \times  \underset{h\times p}{\mathbf{W}^{l1}_n} + \underset{1\times p}{\mathbf{B}^{l1}_n}
)
\times \underset{p\times h}{\mathbf{W}^{l2}_n}
\right )
+
\left(
\mathrm{ReLU} (\underset{s\times h}{\mathrm{X}} \times \underset{h\times (\hat{p}-p)}{\mathbf{M}^{Wl1}_n} +\underset{1\times (\hat{p}-p)}{\mathbf{M}^{bl1}_n}
) 
\times \underset{(\hat{p}-p)\times h}{\mathbf{0}}
\right )
    \\
    \\
&= 
\mathrm{ReLU} (
\underset{s\times h}{\mathrm{X}} \times  \underset{h\times p}{\mathbf{W}^{l1}_n} + \underset{1\times p}{\mathbf{B}^{l1}_n}
)
\times \underset{p\times h}{\mathbf{W}^{l2}_n}
\stepcounter{equation}\tag{\theequation}\label{proof:mlpexpansion}
\end{align*}
\end{proof}

Note that it is not necessary to impose any constraints on the values of $\underset{h\times (\hat{p}-p)}{\mathbf{M}^{Wl1}_n}$ and $\underset{1\times (\hat{p}-p)}{\mathbf{m}^{bl1}_n}$ to achieve function preservation property. Thus, these two matrices can be initialized arbitrarily.

\subsection{Head addition}\label{app:headaddition}

\begin{proof}
\begin{align*}
&\left [ \underset{s\times v}{\mathrm{H}_1} \cdots \ \underset{s\times v}{\mathrm{H}_{(E+1)}} \right ] \times \underset{((E+1)\cdot v) \times h}{\mathbf{\hat{W}}^{O}_n}
\\
\\
&= 
\left [ \underset{s\times v}{\mathrm{H}_1} \cdots \ \underset{s\times v}{\mathrm{H}_{(E+1)}} \right ]
\times
\left [ 
    \begin{array}{c}
    \underset{(E\cdot v) \times h}{\mathbf{W}^{O}_n}
    \\
    \\
    \underset{v\times h}{\mathbf{0}}
    \end{array}
    \right ]
\\
\\
&= 
\left [
\left [ \underset{s\times v}{\mathrm{H}_1} \cdots \ \underset{s\times v}{\mathrm{H}_E} \right ] 
\ 
\underset{s\times v}{\mathrm{H}_{(E+1)}} \right ]
\times
\left [
    \begin{array}{c}
    \underset{(E\cdot v) \times h}{\mathbf{W}^{O}_n}
    \\
    \\
    \underset{v\times h}{\mathbf{0}}
    \end{array}
    \right ]
\\
\\
&= 
\left (
\left [ \underset{s\times v}{\mathrm{H}_1} \cdots \ \underset{s\times v}{\mathrm{H}_E} \right ] 
\times
\underset{(E\cdot v) \times h}{\mathbf{W}^{O}_n}
\right )
+ 
\left (
\underset{s\times v}{\mathrm{H}_{(E+1)}}
\times
\underset{v\times h}{\mathbf{0}}
\right )
\\
\\
&=
\left [ \underset{s\times v}{\mathrm{H}_1} \cdots \ \underset{s\times v}{\mathrm{H}_E} \right ] 
\times
\underset{(E\cdot v) \times h}{\mathbf{W}^{O}_n}
\stepcounter{equation}\tag{\theequation}\label{proof:headaddition}
\end{align*}
\end{proof}

\subsection{Heads expansion} \label{app:headsexpansion}

\begin{proof}
\begin{equation}
\underset{s\times s}{\mathrm{S}_{n,e}}
 :=
 \mathrm{Softmax} \left (
 \frac{1}{\sqrt{k}}  \cdot 
 (
 \underset{s\times h}{\mathrm{X}}\!\!\times\!\underset{h\times k}{\mathbf{W}^{Q}_{n,e}}
 )
 \times
 (
 \underset{s\times h}{\mathrm{X}}\!\!\times\!\underset{h\times k}{\mathbf{W}^{K}_{n,e}}
 )^\top
 \right )
\end{equation}

$\implies$

\begin{align*}
\underset{s\times \hat{v}}{\mathrm{\hat{H}}_e} &=
    \mathrm{Attention}(
\underset{s\times h}{\mathrm{X}}\!\!\times\!\underset{h\times k}{\mathbf{W}^{Q}_{n,e}},
\underset{s\times h}{\mathrm{X}}\!\!\times\!\underset{h\times k}{\mathbf{W}^{K}_{n,e}},
\underset{s\times h}{\mathrm{X}}\!\!\times\!\underset{h\times \hat{v}}{\mathbf{\hat{W}}^{V}_{n,e}}
)
\\
\\
&=
 \underset{s\times s}{\mathrm{S}_{n,e}}
 \times
 \left (
 \underset{s\times h}{\mathrm{X}}\!\!\times\!\underset{h\times \hat{v}}{\mathbf{\hat{W}}^{V}_{n,e}}
 \right )
\\
\\
&= 
 \underset{s\times s}{\mathrm{S}_{n,e}}
 \times
 \left (
 \underset{s\times h}{\mathrm{X}}\!\!\times\!\left [
\underset{h\times v}{\mathbf{W}^{V}_{n,e}}
\ \ \ 
\underset{h\times (\hat{v} - v)}{\mathbf{M}^{W\!V}_{n,e}}
\right ]
\right )
\\
\\
&= 
 \underset{s\times s}{\mathrm{S}_{n,e}}
 \times
 \left [
 \underset{s\times h}{\mathrm{X}}\!\!\times\!\underset{h\times v}{\mathbf{W}^{V}_{n,e}}
\ \ \ 
 \underset{s\times h}{\mathrm{X}}\!\!\times\!\underset{h\times (\hat{v} - v)}{\mathbf{M}^{W\!V}_{n,e}}
\right ]
\\
\\
&=
 \left [
\underset{s\times s}{\mathrm{S}_{n,e}}
 \times
 (
 \underset{s\times h}{\mathrm{X}}\!\!\times\!\underset{h\times v}{\mathbf{W}^{V}_{n,e}}
 )
\ \ \ \ \ 
 \underset{s\times s}{\mathrm{S}_{n,e}}
 \times
 (
 \underset{s\times h}{\mathrm{X}}\!\!\times\!\underset{h\times (\hat{v} - v)}{\mathbf{M}^{W\!V}_{n,e}}
 )
\right ]
\\
\\
&=
 \left [
\underset{s\times v}{\mathrm{H}_e}
\ \ \ \ \ 
 \underset{s\times s}{\mathrm{S}_{n,e}}
 \times
 (
 \underset{s\times h}{\mathrm{X}}\!\!\times\!\underset{h\times (\hat{v} - v)}{\mathbf{M}^{W\!V}_{n,e}}
 )
\right ] \stepcounter{equation}\tag{\theequation}
\end{align*}

$\implies$

\begin{align*}
&\left [ \underset{s\times \hat{v}}{\mathrm{\hat{H}}_1} \cdots \ \underset{s\times \hat{v}}{\mathrm{\hat{H}}_{E}} \right ] \times \underset{(E\cdot \hat{v}) \times h}{\mathbf{\hat{W}}^{O}_n}
\\
&=\left [ \cdots \underset{s\times \hat{v}}{\mathrm{\hat{H}}_e} \cdots
\ \ \vert \ \ e\in [1, E] 
\right ]
\times
\left [
    \begin{array}{c}
    \vdots
    \\
    \underset{v \times h}{\mathbf{\hat{W}}^{O}_{n,e}}
    \\
    \vdots
    \end{array}
    \begin{array}{r}
     \ \vert \ \ e\in [1, E]
    %  \ \forall \ \ e\in [1, E]
     \end{array}
\right ]
\\
\\
&=
\left [ \cdots 
\underset{s\times \hat{v}}{\mathrm{\hat{H}}_e}
\times
\underset{v \times h}{\mathbf{\hat{W}}^{O}_{n,e}}
\cdots
\ \ \vert \ \ e\in [1, E] 
\right ]
\\
\\
&=
\left [ \cdots 
\underset{s\times \hat{v}}{\mathrm{\hat{H}}_e}
\times
\left [ 
    \begin{array}{c}
    \underset{v \times h}{\mathbf{W}^{O}_{n,e}}
    \\
    \\
    \underset{(\hat{v}-v)\times h}{\mathbf{0}}
    \end{array}
\right ]
\cdots
\ \ \vert \ \ e\in [1, E] 
\right ]
\\
\\
&=
\left [ \cdots 
 \left [
\underset{s\times v}{\mathrm{H}_e}
\ \ \ \ \ 
 \underset{s\times s}{\mathrm{S}_{n,e}}
 \times
 (
 \underset{s\times h}{\mathrm{X}}\!\!\times\!\underset{h\times (\hat{v} - v)}{\mathbf{M}^{W\!V}_{n,e}}
 )
\right ]
\times
\left [ 
    \begin{array}{c}
    \underset{v \times h}{\mathbf{W}^{O}_{n,e}}
    \\
    \\
    \underset{(\hat{v}-v)\times h}{\mathbf{0}}
    \end{array}
\right ]
\cdots
\ \ \vert \ \ e\in [1, E] 
\right ]
\\
\\
&=
\left [ \cdots 
 \left [
\underset{s\times v}{\mathrm{H}_e}
\times
    \underset{v \times h}{\mathbf{W}^{O}_{n,e}}
+
 \underset{s\times s}{\mathrm{S}_{n,e}}
 \times
 (
 \underset{s\times h}{\mathrm{X}}\!\!\times\!\underset{h\times (\hat{v} - v)}{\mathbf{M}^{W\!V}_{n,e}}
 )
 \times
    \underset{(\hat{v}-v)\times h}{\mathbf{0}}
\right ]
\cdots
\ \ \vert \ \ e\in [1, E] 
\right ]
\\
\\
&=
\left [ \cdots 
 \left [
\underset{s\times v}{\mathrm{H}_e}
\times
    \underset{v \times h}{\mathbf{W}^{O}_{n,e}}
+
    \underset{s\times h}{\mathbf{0}}
\right ]
\cdots
\ \ \vert \ \ e\in [1, E] 
\right ]
\\
\\
&=
\left [ \cdots 
\underset{s\times v}{\mathrm{H}_e}
\times
\underset{v \times h}{\mathbf{W}^{O}_{n,e}}
\cdots
\ \ \vert \ \ e\in [1, E] 
\right ]
\\
\\
&=
\left [ \cdots \underset{s\times v}{\mathrm{H}_e} \cdots
\ \ \vert \ \ e\in [1, E] 
\right ]
\times
\left [
    \begin{array}{c}
    \vdots
    \\
    \underset{v \times h}{\mathbf{W}^{O}_{n,e}}
    \\
    \vdots
    \end{array}
    \begin{array}{r}
     \ \vert \ \ e\in [1, E]
    %  \ \forall \ \ e\in [1, E]
     \end{array}
\right ]
\\
\\
&=
\left [ \underset{s\times v}{\mathrm{H}_1} \cdots \ \underset{s\times v}{\mathrm{H}_E} \right ] \times \underset{(E\cdot v) \times h}{\mathbf{W}^{O}_n} \stepcounter{equation}\tag{\theequation}
\end{align*}
\end{proof}

\subsection{Attention expansion} \label{app:attnexpansion}

\begin{proof}
% \begin{equation}
%\begin{array}{l}
\begin{align*}
 & \frac{1}{\sqrt{\hat{k}}}
 \cdot
 (
 \underset{s\times h}{\mathrm{X}}\!\!\times\!\underset{h\times \hat{k}}{\mathbf{\hat{W}}^{Q}_{n,e}}
 )
 \times
 (
 \underset{s\times h}{\mathrm{X}}\!\!\times\!\underset{h\times \hat{k}}{\mathbf{\hat{W}}^{K}_{n,e}}
 )^\top 
 \\
 \\
 & {} =
  \frac{1}{\sqrt{\hat{k}}}
 \cdot
 \left (
 \underset{s\times h}{\mathrm{X}}\!\!\times\!
    \left [
    \underset{h\times k}{\mathbf{W}^{Q}_{n,e}}
    \ \ \ 
    \underset{h\times (\hat{k} - k)}{\mathbf{M}^{W\!Q}_{n,e}}
    \right ]
 \right )
 \times
 \left (
 \underset{s\times h}{\mathrm{X}}\!\!\times\!
    \left [
     \frac{\sqrt{\hat{k}}}{\sqrt{k}}
     \cdot 
     \underset{h\times k}{\mathbf{W}^{K}_{n,e}}
    \ \ \ 
    \underset{h\times (\hat{k} - k)}{\mathbf{0}}
    \right ]
 \right )^\top 
 \\
 \\
 & =
  \frac{1}{\sqrt{\hat{k}}}
 \cdot
    \left [
        \underset{s\times h}{\mathrm{X}}\!\!\times\!
        \underset{h\times k}{\mathbf{W}^{Q}_{n,e}}
        \ \ \ 
        \underset{s\times h}{\mathrm{X}}\!\!\times\!
        \underset{h\times (\hat{k} - k)}{\mathbf{M}^{W\!Q}_{n,e}}
    \right ]
 \times
    \left [
      \frac{\sqrt{\hat{k}}}{\sqrt{k}}
      \cdot
      \underset{s\times h}{\mathrm{X}}\!\!\times\!
      \underset{h\times k}{\mathbf{W}^{K}_{n,e}}
      \ \ \ 
      \underset{s\times h}{\mathrm{X}}\!\!\times\!
      \underset{h\times (\hat{k} - k)}{\mathbf{0}}
    \right ]^\top
 \\
 \\
 & =
  \frac{1}{\sqrt{\hat{k}}}
 \cdot
    \left [
        \underset{s\times h}{\mathrm{X}}\!\!\times\!
        \underset{h\times k}{\mathbf{W}^{Q}_{n,e}}
        \ \ \ 
        \underset{s\times h}{\mathrm{X}}\!\!\times\!
        \underset{h\times (\hat{k} - k)}{\mathbf{M}^{W\!Q}_{n,e}}
    \right ]
 \times
    \left [
      \frac{\sqrt{\hat{k}}}{\sqrt{k}}
      \cdot
      \underset{s\times h}{\mathrm{X}}\!\!\times\!
      \underset{h\times k}{\mathbf{W}^{K}_{n,e}}
      \ \ \ 
      \underset{s\times (\hat{k} - k)}{\mathbf{0}}
    \right ]^\top 
 \\
 \\
 & =
  \frac{1}{\sqrt{\hat{k}}}
  \cdot
  \frac{\sqrt{\hat{k}}}{\sqrt{k}}
 \cdot
    \left [
        \underset{s\times h}{\mathrm{X}}\!\!\times\!
        \underset{h\times k}{\mathbf{W}^{Q}_{n,e}}
        \ \ \ 
        \underset{s\times h}{\mathrm{X}}\!\!\times\!
        \underset{h\times (\hat{k} - k)}{\mathbf{M}^{W\!Q}_{n,e}}
    \right ]
 \times
    \left [
      \underset{s\times h}{\mathrm{X}}\!\!\times\!
      \underset{h\times k}{\mathbf{W}^{K}_{n,e}}
      \ \ \ 
      \underset{s\times (\hat{k} - k)}{\mathbf{0}}
    \right ]^\top
 \\
 \\
 & =
 \frac{1}{\sqrt{k}}
 \cdot
    \left [
        \underset{s\times h}{\mathrm{X}}\!\!\times\!
        \underset{h\times k}{\mathbf{W}^{Q}_{n,e}}
        \ \ \ 
        \underset{s\times h}{\mathrm{X}}\!\!\times\!
        \underset{h\times (\hat{k} - k)}{\mathbf{M}^{W\!Q}_{n,e}}
    \right ]
 \times
    \left [
      \underset{s\times h}{\mathrm{X}}\!\!\times\!
      \underset{h\times k}{\mathbf{W}^{K}_{n,e}}
      \ \ \ 
      \underset{s\times (\hat{k} - k)}{\mathbf{0}}
    \right ]^\top
 \\
 \\
 & =
 \frac{1}{\sqrt{k}}
 \cdot
    \left [
        \underset{s\times h}{\mathrm{X}}\!\!\times\!
        \underset{h\times k}{\mathbf{W}^{Q}_{n,e}}
        \ \ \ 
        \underset{s\times h}{\mathrm{X}}\!\!\times\!
        \underset{h\times (\hat{k} - k)}{\mathbf{M}^{W\!Q}_{n,e}}
    \right ]
 \times
    \left [
      \begin{array}{c}
          (
            \underset{s\times h}{\mathrm{X}}\!\!\times\!
            \underset{h\times k}{\mathbf{W}^{K}_{n,e}}
          )^\top
          \\
          \underset{(\hat{k} - k)\times s}{\mathbf{0}}
      \end{array}
    \right ]
 \\
 \\
 & =
 \frac{1}{\sqrt{k}}
 \cdot
    \left (
        (
          \underset{s\times h}{\mathrm{X}}\!\!\times\!
          \underset{h\times k}{\mathbf{W}^{Q}_{n,e}}
         )
        \times
        (
            \underset{s\times h}{\mathrm{X}}\!\!\times\!
            \underset{h\times k}{\mathbf{W}^{K}_{n,e}}
          )^\top
        +
        (\underset{s\times h}{\mathrm{X}}\!\!\times\!
        \underset{h\times (\hat{k} - k)}{\mathbf{M}^{W\!Q}_{n,e}})
        \times
        \underset{(\hat{k} - k)\times s}{\mathbf{0}}
    \right )
 \\
 \\
 & =
 \frac{1}{\sqrt{k}}
 \cdot
    \left (
        (
          \underset{s\times h}{\mathrm{X}}\!\!\times\!
          \underset{h\times k}{\mathbf{W}^{Q}_{n,e}}
         )
        \times
        (
            \underset{s\times h}{\mathrm{X}}\!\!\times\!
            \underset{h\times k}{\mathbf{W}^{K}_{n,e}}
          )^\top
        +
        \underset{s\times s}{\mathbf{0}}
    \right )
 \\
 \\
 & =
 \frac{1}{\sqrt{k}}
 \cdot
        (
          \underset{s\times h}{\mathrm{X}}\!\!\times\!
          \underset{h\times k}{\mathbf{W}^{Q}_{n,e}}
         )
        \times
        (
            \underset{s\times h}{\mathrm{X}}\!\!\times\!
            \underset{h\times k}{\mathbf{W}^{K}_{n,e}}
          )^\top 
          \stepcounter{equation}\tag{\theequation}\label{proof:attentionexpansion}
\end{align*}
%\end{array}
% \end{equation}
\end{proof}

\subsection{Hidden dimension expansion} \label{app:hiddendimexpansion}

\begin{proof}

We demonstrate $\underset{s\times \hat{h}}{\mathrm{\hat{I}}_n} = [\underset{s\times h}{\mathrm{I}_n} \ \ \ \underset{s\times (\hat{h}-h)}{\mathrm{0}}]
\ \forall n\!\in\![0,N]$  by induction on $n$.

Base case $n=0$:
\begin{align*}
\underset{s\times \hat{h}}{\mathrm{\hat{I}}_0} &= 
\underset{s\times h}{\mathrm{\hat{I}}}\!+\!\underset{s\times \hat{h}}{\mathbf{\hat{P}}}
 \\
 \\
 & =
\left [
\underset{s\times h}{\mathrm{I}}
\ \ \ 
\underset{s\times (\hat{h} - h)}{0}
\right ]
+
\left [
\underset{s\times h}{\mathbf{P}}
\ \ \ 
\underset{s\times (\hat{h} - h)}{\mathbf{0}}
\right ]
 \\
 \\
 & =
\left [
\underset{s\times h}{\mathrm{I}}
+
\underset{s\times h}{\mathbf{P}}
\ \ \ 
\underset{s\times (\hat{h} - h)}{0}
\right ].\stepcounter{equation}\tag{\theequation}
\end{align*}

Induction step, assuming $\ \underset{s\times \hat{h}}{\mathrm{\hat{I}}_n} = [\underset{s\times h}{\mathrm{I}_n} \ \ \ \underset{s\times (\hat{h}-h)}{\mathrm{0}}]\ $ holds:

\begin{align*}
      & \mathrm{Norm}_n^\mathrm{MHA}(\underset{s \times h}{\mathrm{\hat{I}}_n}) =
    \biggr [
        \frac{\hat{i}_{\mu,j} \cdot \mathrm{\hat{g}}_{n,j}^\mathrm{MHA}}{\sqrt{\frac{1}{\hat{h}} \sum_{\gamma = 1}^{\hat{h}} (\hat{i}_{\mu,\gamma})^2} } \ \ \ \vert \ \mu \!\in\![1, s] \land j\!\in
        \![1, \hat{h}]  % \ddot{\imath}
    \biggr ]
    \\
    &  =
    \mathrm{Norm}_n^\mathrm{MHA}([\underset{s\times h}{\mathrm{I}_n} \ \ \ \underset{s\times (\hat{h}-h)}{\mathrm{0}}])
    \\
    & =
    \left [
    \biggr [
        \frac{i_{\mu,j} \cdot \mathrm{\hat{g}}_{n,j}^\mathrm{MHA}}{\sqrt{\frac{1}{\hat{h}} \sum_{\gamma = 1}^{\hat{h}} (\hat{i}_{\mu,\gamma})^2} } \ \ \ \vert \ \mu \!\in\![1, s] \land j\!\in
        \![1, h]  % \ddot{\imath}
    \biggr ]
    \biggr [
        \frac{0 \cdot \mathrm{\hat{g}}_{n,j}^\mathrm{MHA}}{\sqrt{\frac{1}{\hat{h}} \sum_{\gamma = 1}^{\hat{h}} (\hat{i}_{\mu,\gamma})^2} } \ \ \ \vert \ \mu \!\in\![1, s] \land j\!\in
        \![h+1, \hat{h}]  % \ddot{\imath}
    \biggr ]
    \right ]
    \\
    & =
    \left [
    \biggr [
        \frac{i_{\mu,j} \cdot \mathrm{\hat{g}}_{n,j}^\mathrm{MHA}}{\sqrt{\frac{1}{\hat{h}} \sum_{\gamma = 1}^{\hat{h}} (\hat{i}_{\mu,\gamma})^2} } \ \ \ \vert \ \mu \!\in\![1, s] \land j\!\in
        \![1, h]  % \ddot{\imath}
    \biggr ]
    \ \ \ \ \ 
    \underset{s\times (\hat{h}-h)}{\mathrm{0}}
    \right ]
    \\
    & =
    \left [
    \biggr [
        \frac{i_{\mu,j} \cdot \mathrm{\hat{g}}_{n,j}^\mathrm{MHA}}{
        \sqrt{
          \frac{1}{\hat{h}}
          (
            \sum_{\gamma = 1}^{h}
              (i_{\mu,\gamma})^2
            +
            \sum_{\gamma = h+1}^{\hat{h}}
              0
          )
        } } \ \ \ \vert \ \mu \!\in\![1, s] \land j\!\in
        \![1, h]  % \ddot{\imath}
    \biggr ]
    \ \ \ \ \ 
    \underset{s\times (\hat{h}-h)}{\mathrm{0}}
    \right ]
    \\
    & =
    \left [
    \biggr [
        \frac{i_{\mu,j} \cdot \mathrm{\hat{g}}_{n,j}^\mathrm{MHA}}{
        \sqrt{
          \frac{1}{\hat{h}}
            \sum_{\gamma = 1}^{h}
              (i_{\mu,\gamma})^2
        } } \ \ \ \vert \ \mu \!\in\![1, s] \land j\!\in
        \![1, h]  % \ddot{\imath}
    \biggr ]
    \ \ \ \ \ 
    \underset{s\times (\hat{h}-h)}{\mathrm{0}}
    \right ]
    \\
    & =
    \left [
    \biggr [
        \frac{i_{\mu,j} \cdot 
        \frac{\sqrt{h}}{\sqrt{\hat{h}}}
        \cdot \mathrm{g}_{n,j}^\mathrm{MHA}}{
        \sqrt{
          \frac{1}{\hat{h}}
            \sum_{\gamma = 1}^{h}
              (i_{\mu,\gamma})^2
        } } \ \ \ \vert \ \mu \!\in\![1, s] \land j\!\in
        \![1, h]  % \ddot{\imath}
    \biggr ]
    \ \ \ \ \ 
    \underset{s\times (\hat{h}-h)}{\mathrm{0}}
    \right ]
    \\
    & =
    \left [
    \biggr [
        \frac{i_{\mu,j} \cdot 
        \mathrm{g}_{n,j}^\mathrm{MHA}}{
        \sqrt{
          \frac{1}{h}
            \sum_{\gamma = 1}^{h}
              (i_{\mu,\gamma})^2
        } } \ \ \ \vert \ \mu \!\in\![1, s] \land j\!\in
        \![1, h]  % \ddot{\imath}
    \biggr ]
    \ \ \ \ \ 
    \underset{s\times (\hat{h}-h)}{\mathrm{0}}
    \right ]
    \\
    & =
    \left [
    \mathrm{Norm}_n^\mathrm{MHA}(\underset{s\times h}{\mathrm{I}_n})
    \ \ \ \ \ 
    \underset{s\times (\hat{h}-h)}{\mathrm{0}}
    \right ]\stepcounter{equation}\tag{\theequation}
\end{align*}
For conciseness, we use the following notation: $\underset{s\times h}{\mathrm{N}_n^c}:=\mathrm{Norm}_n^c(\underset{s\times h}{\mathrm{I}_n})$ and $\underset{s\times \hat{h}}{\mathrm{\hat{N}}_n^c}:= [\underset{s\times h}{\mathrm{N}_n^c} \  \underset{s\times (\hat{h}-h)}{\mathrm{0}}].$

\vspace{8pt}
$\implies$

\begin{align*}
& \underset{s\times \hat{h}}{\mathrm{\hat{I}^{'}}_n} =
\underset{s\times \hat{h}}{\mathrm{\hat{I}}_n} + \mathrm{\hat{MHA}}_n(\underset{s\times \hat{h}}{\mathrm{\hat{N}}_n^\mathrm{MHA}})
\\
& =
 \underset{s\times \hat{h}}{\mathrm{\hat{I}}_n} +
 \left [
   \cdots
      \mathrm{Attention}(
        \underset{s\times \hat{h}}{\mathrm{\hat{N}}_n^\mathrm{MHA}}\!\!\times\!\underset{\hat{h}\times k}{\mathbf{\hat{W}}^{Q}_{n,e}},
        \underset{s\times \hat{h}}{\mathrm{\hat{N}}_n^\mathrm{MHA}}\!\!\times\!\underset{\hat{h}\times k}{\mathbf{\hat{W}}^{K}_{n,e}},
        \underset{s\times \hat{h}}{\mathrm{\hat{N}}_n^\mathrm{MHA}}\!\!\times\!\underset{\hat{h}\times v}{\mathbf{\hat{W}}^{V}_{n,e}}
      )
   \cdots
   \ \vert\ \forall e\!\in\![1, E]
 \right ] \times \underset{(E\cdot v) \times \hat{h}}{\mathbf{\hat{W}}^{O}_n}
\\
&=\!\!
 \underset{s\times \hat{h}}{\mathrm{\hat{I}}_n}\!\!+\!\!
 \left [
   \cdots
      \mathrm{Attention}(
        [\underset{s\times h}{\mathrm{N}_n^\mathrm{MHA}} \  \underset{s\times (\hat{h}-h)}{\mathrm{0}}]\!\!\times\!
            \left [
            \begin{array}{c}
            \underset{h\times v}{\mathbf{W}^{Q}_{n,e}}
              \\
              \\
            \underset{(\hat{h} - h)\times v}{\mathbf{M}^{W\!Q}_{n,e}}
            \end{array}
            \right ]
        ,
        \underset{s\times \hat{h}}{\mathrm{\hat{N}}_n^\mathrm{MHA}}\!\!\times\!\underset{\hat{h}\times k}{\mathbf{\hat{W}}^{K}_{n,e}},
        \underset{s\times \hat{h}}{\mathrm{\hat{N}}_n^\mathrm{MHA}}\!\!\times\!\underset{\hat{h}\times v}{\mathbf{\hat{W}}^{V}_{n,e}}
      )
   \cdots
   \ \vert\ \forall e\!\in\![1, E]
 \right ]
 \!\!\times\!\!\underset{(E\cdot v) \times \hat{h}}{\mathbf{\hat{W}}^{O}_n}
\\
& =
 \underset{s\times \hat{h}}{\mathrm{\hat{I}}_n} +
 \left [
   \cdots
      \mathrm{Attention}(
        \underset{s\times h}{\mathrm{N}_n^\mathrm{MHA}}\!\!\times\!\underset{h\times k}{\mathbf{W}^{Q}_{n,e}},
        \underset{s\times h}{\mathrm{N}_n^\mathrm{MHA}}\!\!\times\!\underset{h\times k}{\mathbf{W}^{K}_{n,e}},
        \underset{s\times h}{\mathrm{N}_n^\mathrm{MHA}}\!\!\times\!\underset{h\times v}{\mathbf{W}^{V}_{n,e}}
      )
   \cdots
   \ \vert\ \forall e\!\in\![1, E]
 \right ] \times \underset{(E\cdot v) \times \hat{h}}{\mathbf{\hat{W}}^{O}_n}
\\
& =
 \underset{s\times \hat{h}}{\mathrm{\hat{I}}_n} +
 \left [
   \cdots
      \underset{s\times v}{\mathrm{H}_{e}}
   \cdots
   \ \vert\ \forall e\!\in\![1, E]
 \right ] \times \left [
\underset{(E\cdot v) \times h}{\mathbf{W}^{O}_n}
\ \ \ 
\underset{(E\cdot v)\times (\hat{h} - h)}{\mathbf{0}}
\right ]
\\
& =
 \underset{s\times \hat{h}}{\mathrm{\hat{I}}_n} +
\left [
 \mathrm{MHA}_n(\underset{s\times h}{\mathrm{N}_n^\mathrm{MHA}})
\ \ \ 
\underset{s\times (\hat{h} - h)}{0}
\right ]
\\
& =
\left [
\underset{s\times h}{\mathrm{I}_n}
\ \ \ \ \ 
\underset{s\times (\hat{h}-h)}{\mathrm{0}}
\right ] +
\left [
 \mathrm{MHA}_n(\underset{s\times h}{\mathrm{N}_n^\mathrm{MHA}})
\ \ \ 
\underset{s\times (\hat{h} - h)}{0}
\right ]
\\
& =
\left [
\underset{s\times h}{\mathrm{I}_n} + \mathrm{MHA}_n(\underset{s\times h}{\mathrm{N}_n^\mathrm{MHA}})
\ \ \ \ \ 
\underset{s\times (\hat{h}-h)}{\mathrm{0}}
\right ]
\\
& =
\left [
\underset{s\times h}{\mathrm{I^{'}}_n}
\ \ \ \ \ 
\underset{s\times (\hat{h}-h)}{\mathrm{0}}
\right ]\stepcounter{equation}\tag{\theequation}
\end{align*}

\vspace{8pt}
$\implies$

Following the demonstration provided for $\mathrm{\hat{Norm}}_n^\mathrm{MHA}(\cdot)$:
\begin{align*}
      & \mathrm{\hat{Norm}}_n^\mathrm{MLP}(\underset{s \times h}{\mathrm{\hat{I}}_n})
      =
    \left [
    \mathrm{Norm}_n^\mathrm{MLP}(\underset{s\times h}{\mathrm{\hat{I}^{'}}_n})\stepcounter{equation}\tag{\theequation}
    \ \ \ \ \ 
    \underset{s\times (\hat{h}-h)}{\mathrm{0}}
    \right ]
    \\
    \\
    & \underset{s\times \hat{h}}{\mathrm{\hat{N}}_n^\mathrm{MLP}} := \mathrm{\hat{Norm}}_n^\mathrm{MLP}(\underset{s \times h}{\mathrm{\hat{I}}_n})\stepcounter{equation}\tag{\theequation}
\end{align*}

\vspace{8pt}
$\implies$

\begin{align*}
\underset{s\times \hat{h}}{\mathrm{\hat{I}}_{n+1}} &=
\mathrm{\hat{TransformerLayer}}_n(\underset{s\times \hat{h}}{\mathrm{\hat{I}}_n})
\\
&=
\underset{s\times \hat{h}}{\mathrm{\hat{I}^{'}}_n} + \mathrm{\hat{MLP}}_n(\underset{s\times \hat{h}}{\mathrm{\hat{N}}_n^\mathrm{MLP}})
\\
&=
\underset{s\times \hat{h}}{\mathrm{\hat{I}^{'}}_n} + \mathrm{\hat{MLP}}_n(\underset{s\times \hat{h}}{\mathrm{\hat{N}}_n^\mathrm{MLP}})
\\
&=
\underset{s\times \hat{h}}{\mathrm{\hat{I}^{'}}_n} +
\mathrm{ReLU}(
\underset{s\times \hat{h}}{\mathrm{\hat{N}}_n^\mathrm{MLP}}
\times
\underset{\hat{h}\times p}{\mathbf{\hat{W}}^{l1}_n}
+
\underset{s\times p}{\mathbf{B}^{l1}_n})
\times
\underset{p\times \hat{h}}{\mathbf{\hat{W}}^{l2}_n}
+
\underset{s\times \hat{h}}{\mathbf{\hat{B}}^{l2}_n}
\\
&=
\underset{s\times \hat{h}}{\mathrm{\hat{I}^{'}}_n} +
\mathrm{ReLU}
(
    [
      \underset{s\times h}{\mathrm{N}_n^\mathrm{MLP}}
      \ \ \ 
      \underset{s\times (\hat{h}-h)}{\mathrm{0}}
    ]
    \times
    \left[
    \begin{array}{c}
    \underset{h\times p}{\mathbf{W}^{l1}_n}
      \\
      \\
    \underset{(\hat{h} - h)\times p}{\mathbf{M}^{Wl1}}
    \end{array}
    \right ]
    +
    \underset{s\times p}{\mathbf{B}^{l1}_n}
)
\times
\underset{p\times \hat{h}}{\mathbf{\hat{W}}^{l2}_n}
+
\underset{s\times \hat{h}}{\mathbf{\hat{B}}^{l2}_n}
\\
&=
\underset{s\times \hat{h}}{\mathrm{\hat{I}^{'}}_n} +
\mathrm{ReLU}
(
    \underset{s\times h}{\mathrm{N}_n^\mathrm{MLP}}
    \times
    \underset{h\times p}{\mathbf{W}^{l1}_n}
    +
    \underset{s\times p}{\mathbf{B}^{l1}_n}
)
\times
\underset{p\times \hat{h}}{\mathbf{\hat{W}}^{l2}_n}
+
\underset{s\times \hat{h}}{\mathbf{\hat{B}}^{l2}_n}
\\
&=
\underset{s\times \hat{h}}{\mathrm{\hat{I}^{'}}_n} +
\mathrm{ReLU}
(
    \underset{s\times h}{\mathrm{N}_n^\mathrm{MLP}}
    \times
    \underset{h\times p}{\mathbf{W}^{l1}_n}
    +
    \underset{s\times p}{\mathbf{B}^{l1}_n}
)
\times
\left [
\underset{p\times h}{\mathbf{W}^{l2}_n}
\ \ \
\underset{p \times (\hat{h} - h)}{\mathbf{0}}
\right ]
+
\left [
\underset{s\times h}{\mathbf{B}^{l2}_n}
\ \ \
\underset{s \times (\hat{h} - h)}{\mathbf{0}}
\right ]
\\
&=
\underset{s\times \hat{h}}{\mathrm{\hat{I}^{'}}_n} +
\left [
\mathrm{ReLU}
(
    \underset{s\times h}{\mathrm{N}_n^\mathrm{MLP}}
    \times
    \underset{h\times p}{\mathbf{W}^{l1}_n}
    +
    \underset{s\times p}{\mathbf{B}^{l1}_n}
)
\times
\underset{p\times h}{\mathbf{W}^{l2}_n}
\ \ \ \ \ 
\underset{s \times (\hat{h} - h)}{\mathbf{0}}
\right ]
+
\left [
\underset{s\times h}{\mathbf{B}^{l2}_n}
\ \ \ \ \ \ 
\underset{s \times (\hat{h} - h)}{\mathbf{0}}
\right ]
\\
&=
\underset{s\times \hat{h}}{\mathrm{\hat{I}^{'}}_n} +
\left [
\mathrm{ReLU}
(
    \underset{s\times h}{\mathrm{N}_n^\mathrm{MLP}}
    \times
    \underset{h\times p}{\mathbf{W}^{l1}_n}
    +
    \underset{s\times p}{\mathbf{B}^{l1}_n}
)
\times
\underset{p\times h}{\mathbf{W}^{l2}_n}
+
\underset{s\times h}{\mathbf{B}^{l2}_n}
\ \ \
\underset{s \times (\hat{h} - h)}{\mathbf{0}}
\right ]
\\
&=
\underset{s\times \hat{h}}{\mathrm{\hat{I}^{'}}_n} +
\left [
\mathrm{MLP}_n(\underset{s\times h}{\mathrm{N}_n^\mathrm{MLP}})
\ \ \ \ \ 
\underset{s \times (\hat{h} - h)}{0}
\right ]
\\
&=
\left [
\underset{s\times h}{\mathrm{I^{'}}_n}
 +
\mathrm{MLP}_n(\underset{s\times h}{\mathrm{N}_n^\mathrm{MLP}})
\ \ \ \ \ 
\underset{s \times (\hat{h} - h)}{0}
\right ]
\\
&=
\left [
\mathrm{\hat{TransformerLayer}}_n(\underset{s\times h}{\mathrm{I}_n}) 
\ \ \ \ \ 
\underset{s \times (\hat{h} - h)}{0}
\right ]
\\
&=
\left [
\underset{s\times h}{\mathrm{I}_{n+1}}
\ \ \ \ \ 
\underset{s \times (\hat{h} - h)}{0}
\right ]\stepcounter{equation}\tag{\theequation}
\end{align*}

Having demonstrated that, after applying the \emph{hidden dimension expansion}:
\begin{equation}
\underset{s\times \hat{h}}{\mathrm{\hat{I}}_{n+1}} = \left [
\underset{s\times h}{\mathrm{I}_{n+1}}
\ \ \ \ \ 
\underset{s \times (\hat{h} - h)}{0}
\right ] \ \forall n\!\in\![1, N+1]
\end{equation} 

The output equivalence can be proven as follows:
\begin{align*}
&\mathrm{\hat{TransformerArchitecture}}(\underset{s\times \hat{h}}{\mathrm{\hat{I}}})
=
\mathrm{\hat{TransformerLayer}}^{\circ N}(\underset{s\times \hat{h}}{\mathrm{\hat{I}}}\!+\!\underset{s\times \hat{h}}{\mathbf{\hat{P}}})\ \times \underset{\hat{h}\times o}{\mathbf{\hat{W}}^{out}}
\\
&=
\underset{s\times \hat{h}}{\mathrm{\hat{I}}_{N+1}}
\times \underset{\hat{h}\times o}{\mathbf{\hat{W}}^{out}}
=
\left [
\underset{s\times h}{\mathrm{I}_{N+1}}
\ \ \ \ \ 
\underset{s \times (\hat{h} - h)}{0}
\right ]
\times
\left [
\begin{array}{c}
\underset{h\times o}{\mathbf{W}^{out}}
  \\
  \\
\underset{(\hat{h} - h)\times o}{\mathbf{M}^{Wout}}
\end{array}
\right ]
=
\underset{s\times h}{\mathrm{I}_{N+1}}
\times \underset{h\times o}{\mathbf{W}^{out}}
\\
&= \mathrm{TransformerArchitecture}(\underset{s\times h}{\mathrm{I}})\stepcounter{equation}\tag{\theequation}
\end{align*}
\end{proof}

\subsection{Layer addition} \label{app:layeraddition}

\begin{proof}

\begin{align*}
&
\mathrm{MHA}_n(\underset{s\times h}{\mathrm{X}_n}) = \left [ \underset{s\times v}{\mathrm{H}_1} \cdots \ \underset{s\times v}{\mathrm{H}_E} \right ] \times \underset{(E\cdot v) \times h}{\mathbf{0}} = \underset{s\times h}{\mathrm{0}}\stepcounter{equation}\tag{\theequation}
\\ \\
&
\mathrm{MLP}_n(\underset{s\times h}{\mathrm{X}_n}) =
\mathrm{ReLU}(
\underset{s\times h}{\mathrm{X}_n} \times \underset{h\times p}{\mathbf{W}^{l1}_n} + \underset{s\times p}{\mathbf{B}^{l1}_n}) \times \underset{p\times h}{\mathbf{0}}
+ \underset{s\times h}{\mathbf{0}} = \underset{s\times h}{\mathrm{0}}\stepcounter{equation}\tag{\theequation}
\\
\\
&
\underset{s\times h}{\mathrm{I^{'}}_n} = \underset{s\times h}{\mathrm{I}_n} + \mathrm{MHA}_n(\mathrm{Norm}_n^\mathrm{MHA}(\underset{s\times h}{\mathrm{I}_n})) = \underset{s\times h}{\mathrm{I}_n} + \underset{s\times h}{\mathrm{0}_n} = \underset{s\times h}{\mathrm{I}_n} \stepcounter{equation}\tag{\theequation}
\\
\\
&
\mathrm{TransformerLayer}_n(\underset{s\times h}{\mathrm{I}_n}) = \underset{s\times h}{\mathrm{I}_n} + \mathrm{MLP}_n(\mathrm{Norm}_n^\mathrm{MLP}(\underset{s\times h}{\mathrm{I}_n}))  = \underset{s\times h}{\mathrm{I}_n} + \underset{s\times h}{\mathrm{0}_n} = \underset{s\times h}{\mathrm{I}_n} \stepcounter{equation}\tag{\theequation}
\end{align*}
\end{proof}

Note that the function preserving property holds even if normalization is applied after the MLP and MHA components as $\mathrm{Norm}(\cdot)$ outputs zeros for zeros input.

\end{document}